\documentclass{article}
\usepackage{fourier}
\usepackage{amsmath,amssymb,amsthm}
\usepackage[margin=1in]{geometry}

\usepackage[numbers,compress]{natbib}

\usepackage{graphicx} 
\graphicspath{{./figures/}}
\usepackage{subfigure}

\usepackage{algorithm}
\usepackage{algorithmic}
\usepackage{color}

\usepackage[utf8]{inputenc} 
\usepackage[T1]{fontenc}    
\usepackage{hyperref}       
\usepackage{url}            
\usepackage{booktabs}       
\usepackage{amsfonts}       
\usepackage{nicefrac}       
\usepackage{microtype}      

\usepackage{wrapfig}
\usepackage{multicol}
\usepackage{stmaryrd}

\newcommand{\mak}{\texttt{max\_active\_keys}}

\title{AMPNet: Asynchronous Model-Parallel Training for Dynamic Neural Networks}

\author{
Alexander L. Gaunt,
Matthew A. Johnson,
Maik Riechert,
Daniel~Tarlow\thanks{Currently at Google Brain},\\
Ryota~Tomioka,
Dimitrios~Vytiniotis,
Sam Webster\vspace{5mm}\\ 
Microsoft Research\\
Cambridge, United Kingdom
}
\begin{document}

\maketitle

\begin{abstract}
New types of machine learning hardware in development and entering the market hold the
promise of revolutionizing deep learning in a manner as profound as GPUs. However, existing software
frameworks and training algorithms for deep learning have yet to evolve to
fully leverage the  capability of the new wave of silicon. We already
see the limitations of existing algorithms for models that exploit structured input
via complex and instance-dependent control flow, which prohibits minibatching. We present an {\em asynchronous  model-parallel} (AMP) training algorithm that is 
specifically motivated by training on networks of interconnected
 devices. 
Through an implementation on multi-core CPUs, we show that AMP training converges to the same
accuracy as conventional synchronous training algorithms in a similar number of epochs, but utilizes 
the available hardware more efficiently even for small minibatch sizes, resulting in
significantly shorter overall training times. Our framework opens the
 door for scaling up a new class of deep learning models that cannot be efficiently trained today.
\end{abstract}

\section{Introduction}
A new category of neural 
networks is emerging whose common trait is their ability to react in dynamic and unique ways to properties of their
input. Networks like tree-structured recursive neural networks \citep{socher2013recursive,tai2015improved} and
graph neural networks (GNNs) \citep{scarselli2009graph,li2015gated,gilmer2017neural} defy the modern GPU-driven paradigm
of minibatch-based data management. Instead, these networks take a tree or a graph as input and carry out a computation that 
depends on their individual structures. We refer to this new class of models with dynamic control flow as 
{\em dynamic neural networks.}

Modern neural network frameworks are certainly capable of expressing dynamic networks. TensorFlow \cite{abadi2016tensorflow} 
introduces \texttt{cond}, \texttt{while\_loop}, and other higher order functional abstractions, while 
Chainer \cite{tokui2015chainer},
DyNet \cite{neubig2017dynet}, and PyTorch \cite{pytorch}
dynamically construct the computation graph using the control flow of the host language. 
However, {\em training} these networks with existing software
frameworks and hardware can be painfully slow because 
these networks require highly irregular, non-uniform computation that
depends on individual instances.
This makes batching impractical or impossible, thus
causing the cost of matrix-vector product to be dominated by the cost of loading the weights from DRAM -- typically
orders of magnitude slower than the peak compute on both CPUs and GPUs 
\footnote{ For example, the TitanX GPU performs $10^{13}$ FLOPS
but only $10^{11}$ floats/s can be brought into the chip due to memory
bandwidth (480 GB/s).}. Moreover, these frameworks are not
typically optimized with single instance batch size in mind. Dynamically unfolding the 
computation graph, for example, is a concern when there are not
enough instances to amortize the cost for it.
\footnote{Recently proposed TensorFlow Fold \cite{looks2017deep} mitigates these issues with dynamic 
batching. (Section \ref{sec:relatedwork})}

With limited batching, we show in this paper that a way to scale up dynamic models is by exploiting an extreme
form of {\em model parallelism}, amenable to distributed execution on a cluster of interconnected
compute devices.
By model parallelism, we not only mean computing disjoint parts of the
computational graph in parallel, but also computing sequential operations in the graph
in a pipeline-parallel fashion \citep[see e.g.,][]{chen2012pipelined}.



\begin{figure}[tb]
 \begin{center}
  \includegraphics[width=1.0\columnwidth]{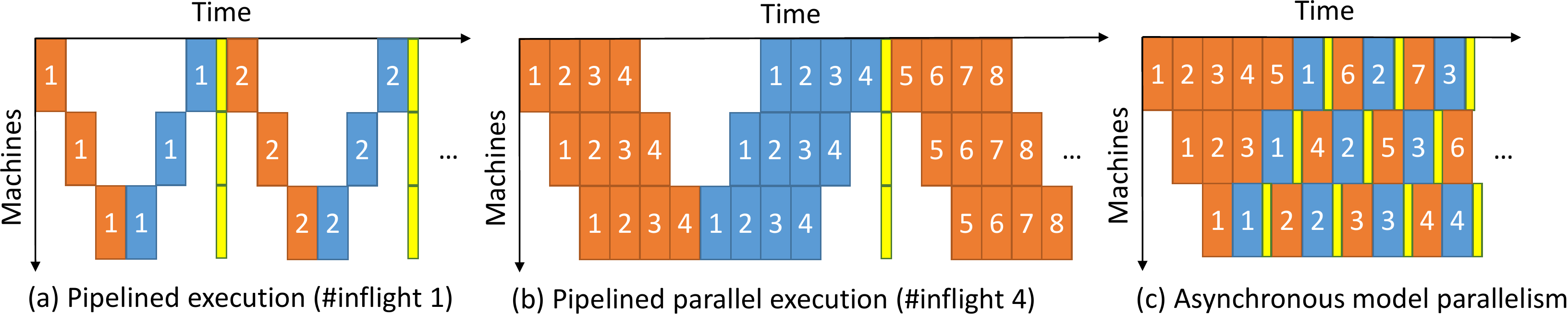}
  \caption{Gantt charts comparing pipelined synchronous model parallelism
  and asynchronous model parallelism. Orange, blue, and yellow boxes
  correspond to forward, backward, and parameter update operations,
  respectively. The numbers in the boxes indicate instance IDs.}
  \label{fig:async}
 \end{center}
\end{figure}

Conventional (pipeline) model parallelism, however, can only maximize
device utilization if we can keep the pipeline full at all times.
Unfortunately, as we show in Figure \ref{fig:async},
keeping a conventional pipeline full is
at odds with convergence speed due to a decreased parameter update
frequency; compare Figure \ref{fig:async} (a) and (b). This is analogous
to the trade-off we face in batching.
To overcome this problem,
we propose {\em asynchronous model-parallel (AMP) training}, where we
allow asynchronous gradient updates to occur, whenever enough gradients have
been accumulated; see Figure \ref{fig:async} (c). With this design we
aim for both high device utilization and update frequency.

In this setting, however, model parameters may be updated between the forward and the backward computation
of an instance, introducing gradient ``staleness''. 
Despite staleness, we show that AMP training can
converge fast with good hardware utilization. Specifically, our contributions are:
\begin{itemize}
\item We present the AMPNet framework for efficient distributed training of dynamic networks.
\item We present an intermediate representation (IR)
  with explicit constructs for branching and joining control flow that supports AMP training.
  Unlike previous work that considers static computation
  graphs for static control flow (e.g., Caffe), and dynamic computation graphs for dynamic control flow (e.g., Chainer), our IR encodes a
  static computation graph to execute dynamic control flow\footnote{Our IR bears similarity to TensorFlow but we discuss differences in Section~\ref{sec:relatedwork}.}.
  As a consequence, training becomes easy to distribute and parallelize. Further, IR nodes can process forward and backward messages from multiple instances
  at the same time and seamlessly support simultaneous training and inference.

\item We show that, thanks to explicit control flow constructs, our IR can readily encode {\em replicas}, a form of {\em data parallelism}
      (see Sec.~\ref{sec:replicas}).
  In addition, our IR includes operators for data aggregation which recover forms of {\em batching}. These features can further improve 
  efficiency, even on CPUs.
\item We show that AMP training converges to similar accuracies as synchronous algorithms but often significantly faster. (Sec.~\ref{sec:experiments})
      For example on the QM9 dataset~\cite{ruddigkeit2012enumeration, ramakrishnan2014quantum}
		our implementation of gated graph sequence neural network (GGSNN)
      \cite{li2015gated} on a 16 core CPU runs 9x faster than
      a (manually optimized) TensorFlow CPU implementation and 2.1x faster
      than a TensorFlow GPU implementation on the TitanX GPU, because it can better exploit sparsity.
      Though we do not aim to compete across-the-board with mature frameworks such as TensorFlow, our
      evaluation proves that AMPNet is particularly beneficial for dynamic networks.
\end{itemize}

In summary, our work demonstrates the benefits of AMP training
and gives a novel way to design and deploy neural network libraries with dynamic control flow.
Together these contributions open up new ways to scale up dynamic networks on interconnected compute devices. 
Inspired by the increasing investment and innovation in custom silicon
for machine learning (i.e., FPGAs \cite{farabet2011large,caulfield2016cloud} and ASICs \cite{jouppi2017datacenter}),
we perform a 
simple calculation on the QM9 dataset that shows that AMPNet
on a network of 1 TFLOPS devices can be 10x faster than our CPU runtime
requiring only 1.2 Gb/s network bandwidth (Sec.~\ref{sec:conclusion}).

\section{Neural networks with complex and dynamic control flow}
\label{sec:models}

    \begin{figure}
\begin{center}
   \includegraphics[scale=0.78]{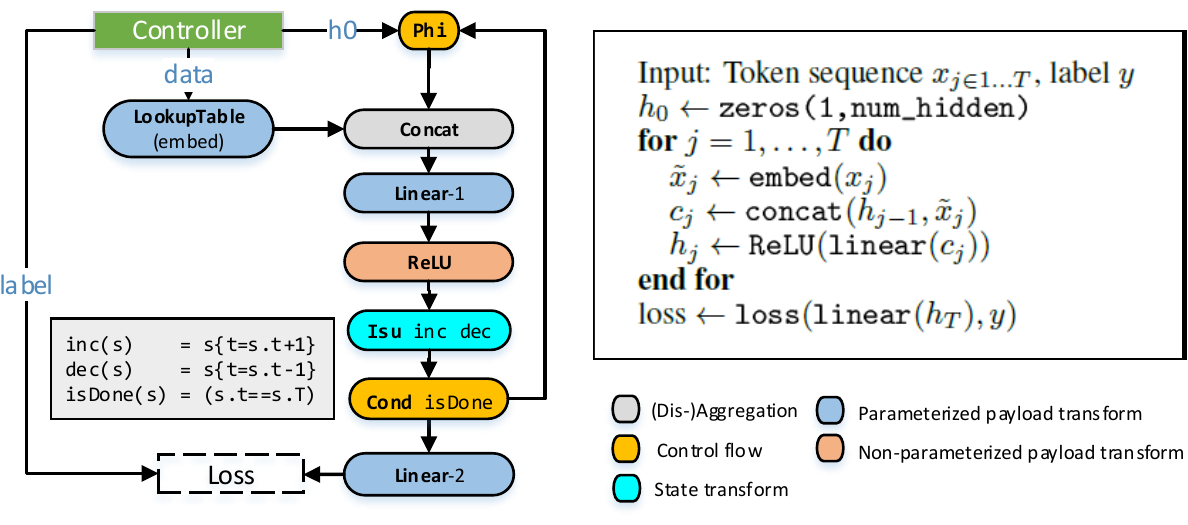}
\end{center}
   \caption{Variable-length RNN in IR and pseudocode (colors denote IR node types)}
  \label{fig:rnn-ctrl}
\end{figure}

Below we highlight three models with dynamic control flow, which will be studied in depth in this paper:

        {\em Variable-length RNNs} iterate over the tokens of variable-length sequences.
        Pseudo-code for a simple RNN is given in Figure~\ref{fig:rnn-ctrl}. The linear layer can be substituted
with a more sophisticated unit such as a gated recurrent unit~\cite{chung2014empirical}.
Though each instance has a different length,
it {\em is} possible to add padding to enable batching. However this may lead to
limited speedup due to variability in sequence lengths.

{\em Tree-structured neural networks} are powerful models used for parsing of
natural language and images, semantic representation, and sentiment
analysis \cite{socher2011parsing,bowman2016fast,socher2013recursive,tai2015improved}. They require evaluation
of (potentially multiple) trees with  shared parameters but different topology for each instance. Even if one only needs to evaluate a single
computational tree per instance as in \cite{socher2013recursive,tai2015improved}, the tree is instance-specific and batching requires nontrivial planning~\cite{looks2017deep}.
A simple form of tree neural network performs a bottom up traversal of the instance,
starting from an embedding of the leaves. At each level the values from the child nodes are
concatenated and sent through a specialized unit (e.g. LSTM). The
result is then propagated further up the tree. 
Backpropagation over the tree structure is known as
backpropagation through structure \cite{goller1996learning}.

{\em Graph neural networks}~\cite{scarselli2009graph,li2015gated,gilmer2017neural} combine both the temporal
recurrence in variable length RNN and recurrence over the structure in
tree RNN. GNNs can be seen as performing aggregation/distribution operations over a general
graph structure with shared parameters. 

Apart from the models above, there exist many recently proposed models with
flexible control flow (e.g. hierarchical memory networks~\cite{chandar2016hierarchical}, neural programmer interpreters~\cite{reed2015neural}, adaptive computation networks~\cite{graves2016adaptive, figurnov2016spatially}, and networks with stochastic depth~\cite{huang2016deep}), to which our framework can be applied.

\section{Asynchronous model-parallel training}
\label{sec:amp}

The basic idea behind AMP training is to distribute a computation graph across compute nodes and communicate
activations. 
For training, the nodes of the computation graph exchange forward or backward messages.
Parameterized computations (e.g. fully-connected layers) can individually accumulate gradients computed from backwards
messages. Once the number of accumulated gradients
since the last update exceeds a threshold {\tt min\_update\_frequency},
a local update is applied and the accumulator gets cleared. 
The local parameter update occurs without further communication or synchronization
with other parameterized computations. The staleness of a gradient can
be measured by the number of updates between the forward and backward
computation that produces the gradient. Small {\tt min\_update\_frequency} may
increase gradient staleness. On the other hand, large {\tt min\_update\_frequency} can reduce the variance
of the gradient but can result in very infrequent updates and also slow down convergence. In addition, \mak{} controls the maximum number
of {\em active instances} that are in-flight at any point in time. By setting $\mak = 1$ we restrict to single-instance processing.\footnote{Note this is usually, but not always, equivalent
  to synchronous training. For example, a single instance can be comprised of a stream of messages (e.g. tree nodes in a tree RNN) and depending on the model some updates may
  occur asynchronously, even if all the messages in-flight belong to a single instance.} 
More in-flight messages generally increase hardware utilization, but may also
increase gradient staleness. 
We have implemented an AMPNet runtime for multi-core CPUs, the details
of which are given in Appendix A.
 Section~\ref{sec:experiments} demonstrates the
effects of these parameters.

\section{A static intermediate representation for dynamic control flow}\label{sec:ir}
\newcommand{\pldtx} {{{\bf\tt pldtx}}}
\newcommand{\ppldtx}{{{\bf\tt ppldtx}}}
\newcommand{\cpldtx}{{{\bf\tt cpldtx}}}
\newcommand{\cpldx} {{{\bf\tt cpldtx}}}
\newcommand{\istx}{{{\bf\tt Isu}}}
\newcommand{\gstx}{{{\bf\tt sttx}}}
\newcommand{\cond}{{{\bf\tt Cond}}}
\newcommand{\irphi}{{{\bf\tt Phi}}}
\newcommand{\group}{{{\bf\tt Group}}}
\newcommand{\ungroup}{{{\bf\tt Ungroup}}}
\newcommand{\bcast}{{{\bf\tt Bcast}}}
\newcommand{\flatmap}{{{\bf\tt Flatmap}}}
\newcommand{\concat}{{{\bf\tt Concat}}}
\newcommand{\irsplit}{{{\bf\tt Split}}}
\newcommand{\msg}[2]{{\tt Msg}\;#1\;#2}

\paragraph{Overview}
Motivated by the need to distribute dynamic networks on networks of interconnected devices and
apply AMP training, we have designed a static graph-like
{\em intermediate representation} (IR) that can serve as a target of compilation for
high-level libraries for dynamic networks (e.g. TensorFlow or our own frontend), and can
itself admit multiple backends (e.g. the multi-core CPU runtime that we consider in detail
in this paper, or a network of accelerators). The key feature of our IR is that it is a static graph,
but can execute dynamic and instance-dependent control flow decisions.

A neural network model is specified by (i) an IR graph,
and (ii) a specialized controller loop that pumps instances and
other data -- e.g. initial hidden states -- and is
responsible for throttling asynchrony.

Each IR node can receive and process either forward messages (from its
predecessor in the IR graph) or backward messages (from its
successors). During training, forward propagation is carried out by
passing forward messages through the IR graph. Each message consists
of a {\em payload} and a {\em state}. The payload is typically
a tensor, whereas the state is typically model-specific and
is used to keep track of algorithm and control flow information. For
example, in a variable-length RNN the state contains the instance
identifier, the current position in the sequence, and the total sequence length for the instance.
The final loss layer initiates the backward propagation through the IR graph.
An invariant of our IR is that for every forward message that is generated by a
node with a specific state, this node will eventually receive a backward message with
the same state. Depending on \mak{} (Section~\ref{sec:amp})
multiple forward or backward 
messages can be in-flight, from one or more instances.

In the rest of this section we discuss the most important IR
nodes along with their operational semantics, and show how they are used
in the example models from the previous section.

\paragraph{Payload transformations}
{\em Parameterized payload transform} (PPT) nodes can be used to encode, for
instance, fully connected layers. They apply a transform in the forward
pass, but also record the activation in order to use it to compute gradients
in the backward pass. An activation is recorded by keying on the state of the
message, and hence this state must include all necessary information to allow the
node to process multiple messages from potentially different instances without
conflating the activations. We require specifications of the
forward and the backward transformation, the operation to produce
a new gradient, as well as the state keying function to be used. A PPT node
may decide to independently apply accumulated gradients to update its parameters.
For transformations that do not involve parameters (e.g. ReLUs) our IR offers a
simpler {\em non-parameterized payload transform}.

\paragraph{Loops, state, and control flow}
A {\em condition} node ($\cond~f$) is parameterized by a function $f$ that
  queries the {\em state} (but not the payload) of the incoming message and,
  based on the response, routes the input to one of the successor nodes.
A {\em join} node ($\irphi$) propagates the messages it
  receives from each of its ancestor nodes but records the
  origin so that in the backward pass it can backpropagate them to
  the correct origin. Like PPT nodes, a $\irphi$ node
  must be parameterized over the keying function on the state of the incoming message.
An {\em invertible state update} node ($\istx~f~f^{-1}$) is parameterized by
  two functions $f$ and $f^{-1}$ that operate on the state of a message, and
  satisfy $f^{-1}(f(x)) = x$.

\begin{figure}
 \begin{center}
  \includegraphics[scale=0.9]{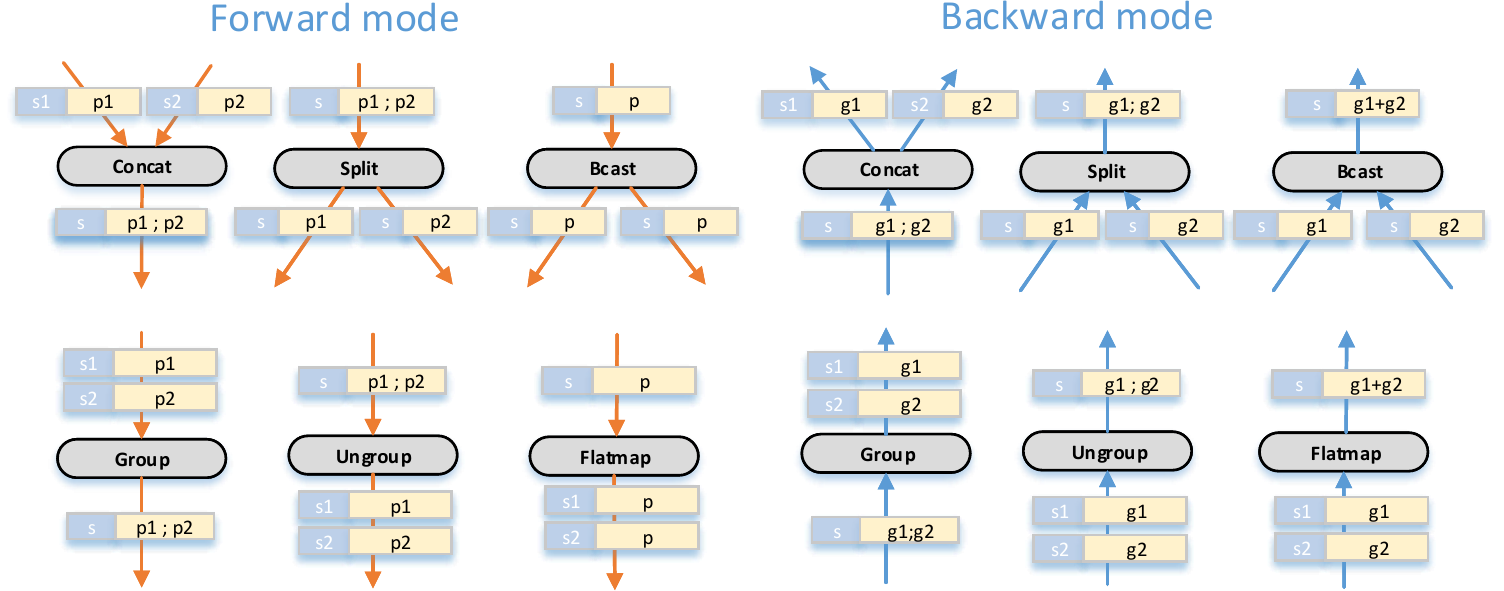}
 \end{center}
   \caption{(Dis-)aggregation combinators (forward mode, left; backward mode, right)} 
  \label{fig:aggregation}
\end{figure}

Figure~\ref{fig:rnn-ctrl} shows how to encode an RNN. The
controller pumps sequence tokens into a lookup table -- just a PPT node, where the parameter is the embedding table and is also being learned.
The controller also pumps labels to the loss layer (dashed boxes are compound graphs whose details
we omit), and an initial hidden state $h_0$ for every
sequence. Message states contain the sequence time-step. Following the embedding,
messages are concatenated (\concat~node, see next paragraph) with the
hidden state, and the result goes into a linear node followed by a ReLU activation.
The $\istx$ node increments the time-step, and the conditional node tests
whether the end of the sequence has been reached. Depending on the answer it either propagates the hidden state back to $\irphi$, or pushes the hidden state
to the final linear and loss layers. In backward mode, the gradient is propagated inside the body of the loop, passes through the $\istx$ (which decrements the time-step), and
reaches the $\irphi$ node. The $\irphi$ node will (based on information from the forward phase) either backpropagate
to the $\cond$ node, or to the controller. Hence the loop is executed in both the forward and backward direction.

\paragraph{Aggregation and disaggregation}

Our IR offers several constructs for aggregation and disagreggation; the most important ones are outlined below, and their behavior is summarized in Figure~\ref{fig:aggregation}.
    $\concat$, $\irsplit$, and $\bcast$ perform concatenation,
partition, and broadcast of incoming messages as their names suggest.
%
   $\group$ can group together several incoming
     messages based on their state. The output
     message contains a tensor composed of the input payloads, whereas the state is a
     function of the incoming states.
          In forward mode
     $\group$ (and also $\concat$) must key on this new state to cache the
     states of the original messages, so as to restore
     those in the backward phase. $\ungroup$ is a symmetric version of $\group$.
     $\flatmap$ creates a sequence of outgoing messages per incoming message, with replicated payload and new states
     given by a state generation function that is a parameter of the node. The node keys on the outgoing states and caches
     the incoming state and number of expected messages, so as to sum all the gradients and restore
     the original state in backward mode.

\begin{figure}[tb]
{\centering 
\subfigure[Gated Graph Sequence Neural Network. RNNCell is a placeholder for a recurrent structure (e.g. GRU, LSTM), the details of which we omit.]{\includegraphics[width=0.48\textwidth,trim={7mm 7mm 7mm 7mm},clip]{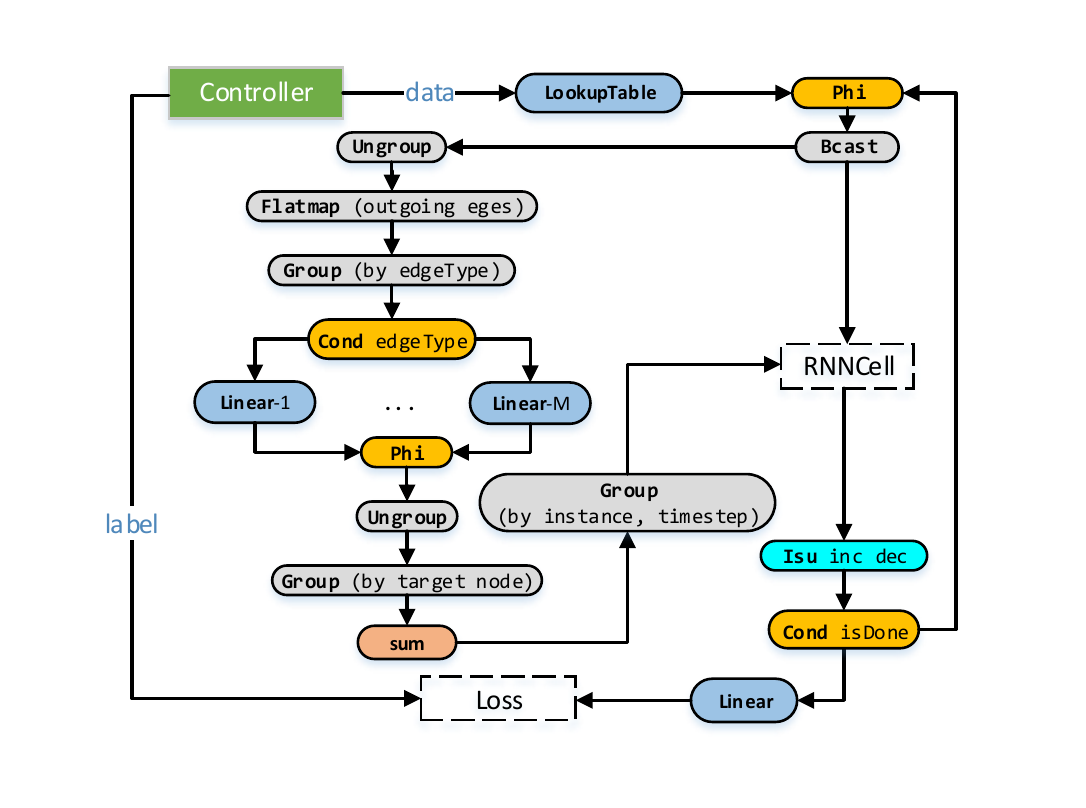}\label{fig:gnn-ctrl}}\hspace{5mm}
\subfigure[IR graph for an RNN with
 replicas.]{\includegraphics[width=.4\textwidth]{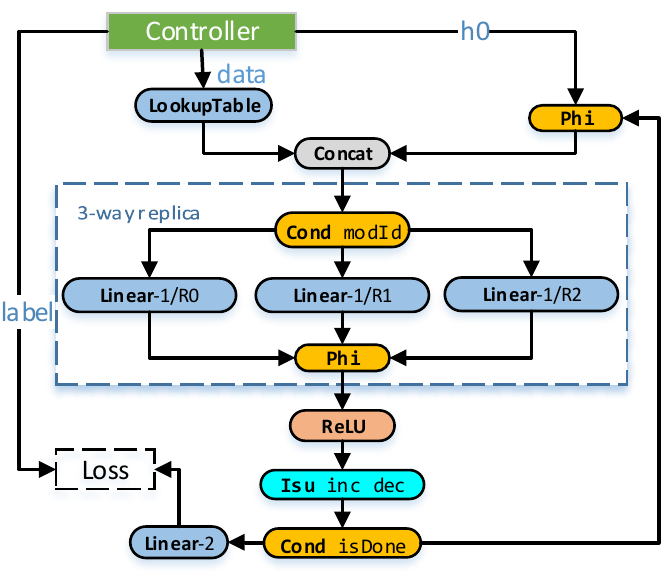}\label{fig:rnn-replicated-ctrl}}
}
 \caption{IR graphs for Gated Graph Sequence Neural Network and RNN
 with replicas.}
 \end{figure}

     Figure~\ref{fig:gnn-ctrl} describes a GNN that combines aggregation on the structure of a graph instance
     with an outer iteration. The iteration controls in effect the locality of information propagation.
The controller pumps data, as before, to a lookup table
and labels for this instance to the loss layer.
The lookup table emits payloads that are matrices where each row corresponds to the embedding
of an instance node, and states that contain the current iteration counter, the instance id, and a reference to the graph structure.
The messages are broadcast and ungrouped so that each outgoing message corresponds to each node of the graph instance.
Next, each message is
goes through a $\flatmap$ node that replicates the payload for each {\em outgoing edge} and creates states that record the incoming node, outgoing node, and type of that edge,
resulting in a stream of messages,  one for each edge in the graph. Next, all edges are grouped by edge type and each group is sent to a designated linear layer.
Each group is then dismantled back and edges are re-grouped by their target node. Each group is passed through a non-parameterized payload transformation that sums together all payloads.
The result is a stream of messages where each message contains an aggregated value for a graph node. Finally, we group back all these aggregated values and send the result to the RNNCell
for another outer iteration. We note that this constitutes a form of batching -- the information about all nodes is batched together before been sent to the RNNCell.

\section{Interaction with data parallelism and replicas}
\label{sec:replicas}
Pipeline-style parallelism can often be augmented with forms of data parallelism.
Consider the RNN in Fig.~\ref{fig:rnn-ctrl}. The only heavy operation (Linear-1) in the body of the loop
is going to act as a bottleneck for computation.
One solution is to split the linear layer into smaller tiles and compute them in parallel.
 This is expressible in our IR but the linear 
 operation needs to be large enough to benefit from tiling in this way.
 Another approach is to replicate the linear
layer in full. Fortunately this requires only minimal new machinery -- we can replicate the
linear layer and place the replicas inside $\cond$ and $\irphi$ nodes as in Figure~\ref{fig:rnn-replicated-ctrl}.
Different instances or messages from the same instance but with different position in the sequence can be processed in an
(pipeline-)parallel fashion using 3 replicas in this case. To enable parameters to be shared among the replicas,
we have implemented infrequent end-of-epoch replica synchronization (averaging) to keep the communication cost negligible, as well as a
message-passing protocol asynchronous trigger of whole-replica group synchronization, but
found that infrequent synchronization was sufficient for fast convergence.

\section{Experiments}
\label{sec:experiments}

We evaluate AMPNet using the dynamic models introduced in
Section~\ref{sec:models}. For completeness, we additionally consider a
simple multi-layer perceptron (MLP) as an example of a static network with instances
that are easy to batch. For each model we select a dataset
and compare the throughput and convergence profile of AMPNet against traditional training schemes implemented in TensorFlow.

\paragraph{MLP: MNIST}
As preliminary task, we train a 4-layer perceptron with ReLUs on
MNIST~\citep{lecun1998mnist}. We choose $784$-dimensional hidden units,
and we affinitize the 3 linear operations on individual workers
(or threads; see Appendix A). Both AMP runtime and TensorFlow use SGD with learning rate 0.1 and batch size of 100.

\paragraph{RNN: List reduction dataset}
As a starting point for experiments on networks with complex control flow
we use a synthetic dataset solved by a vanilla RNN. Specifically, we train
an RNN to perform reduction operations on variable length lists of digits.
Each training instance is a sequence of at most 10 tokens: The first token
indicates which of 4 reduction operations
\footnote{The operations considered in our toy dataset act on a list $L$ and are
expressed in python syntax as: \texttt{mean($L$)}, \texttt{mean($L$[0::2])-mean($L$[1::2])},
 \texttt{max($L$)-min($L$)} and \texttt{len($L$)}.}
is to be performed, and the remaining tokens represent the list
of digits. The output is the result of the calculation
rounded modulo 10.
The dataset consists of $10^5$ training and $10^4$ validation instances.

We present this task as a classification problem to a vanilla RNN with
ReLU activation and a hidden dimension of 128.
All parameterized operations are affinitized on individual workers.
We bucket training instances into batches of 100 sequences (in the baseline and in AMPNet).

\paragraph{Tree-LSTM: Stanford Sentiment Treebank}
As a non-synthetic problem, we consider a real-world sentiment
classification dataset~\cite{socher2013recursive}
consisting of binarized constituency parse trees of English sentences
with sentiment labels at each node. Following Tai et al.
\cite{tai2015improved}, we use 8,544 trees for training,
1,101 trees for validation, and 2,210 trees for testing.

We use a Tree LSTM for this classification task based on the
TensorFlow Fold~\cite{looks2017deep} benchmark model.
Both the AMP and Fold models are trained following \cite{tai2015improved}
with the additional architectural modifications proposed by
\cite{looks2017deep, semeniuta2016recurrent}.
Furthermore, we split our Tree-LSTM cell into Leaf LSTM
and Branch LSTM cells. This does not affect the expressiveness of the
model because the LSTM cell receives either zero input (on branch) or zero hidden
states (on leaves); i.e., the two cells do not share weights except for the bias
parameters, which are learned independently in our implementation.
We compare the time to reach 82 \%  fine
grained (5 classes) accuracy (averaged over all the nodes) on the
validation set.

\paragraph{GNN: Facebook bAbI 15 \& QM9 datasets}
We verify our GNN implementation using a toy logic deduction
benchmark (bAbI task 15 \cite{weston2015towards}) and study a real-world application for GNNs:
prediction of organic molecule properties from structural formulae in the
QM9 dataset~\cite{ruddigkeit2012enumeration, ramakrishnan2014quantum}. GNNs have
previously been applied to these tasks in \cite{li2015gated} and \cite{gilmer2017neural} respectively.

For the bAbI 15 dataset we inflate each graphs from the default 8 nodes to 54 nodes to increase the
computational load, but we preserve the two-hop complexity of the deduction task.
The architecture of the model follows \cite{li2015gated} with a hidden dimension of 5,
and 2 propagation steps.

For the QM9 dataset we concentrate on prediction of the norm of a molecule's
dipole moment using a regression layer build on the propagation model from \cite{li2015gated} (corresponding
to the simplest setting in \cite{gilmer2017neural}). We use a hidden dimension of 100 and 4
propagation steps, initializing the graph nodes (atoms) following \cite{gilmer2017neural}. The molecules
contain up to 29 atoms and in a TensorFlow baseline we bucket molecules into batches
of 20 with atom counts differing by at most 1 within a batch. Following \cite{gilmer2017neural}, we report regression accuracies in multiples of a target accuracy from the chemistry community.

\paragraph{Results}

\begin{table}[tb]
  \begin{center}
   \caption{Time to convergence to target validation accuracy. The time
   to convergence can be broken down into number of epochs and the
   throughput (instances/s). The
   target accuracy is shown inside parentheses next to each dataset.
$\texttt{mak}$ is a short-hand for \mak\;defined in 
 Sec.~\ref{sec:amp}; $\texttt{mak}=1$ corresponds to synchronous
   training for MNIST and minimal asynchrony arising from just one
   in-flight instance for other models with recursive structures. } 
   \label{tab:results}
{\footnotesize
   \begin{tabular}{@{\extracolsep{4pt}}llllllll@{}}
   \hline
   \hline
   & \multicolumn{4}{l}{AMP}
   & \multicolumn{3}{l}{TensorFlow} \\
  \cline{2-5}  \cline{6-8}
    & \texttt{mak}   & time (s) & epochs & inst/s
    & time (s) & epochs & inst/s \\
    \hline
    MNIST (97\%) & 1 & 130       & 4  & 1949 &\\
    & 4 & 44 (3x) &  4 &  5750
    & {\bf 34.5} & 3 & 5880  \\
 \hline
    List reduction (97\%)      & 1 & 82.9 & 9 & 12k \\
    & 4 & 69.7 (1.2x) & 9 & 14k\\
    & 16 & 64.9 (1.3x)& 9 & 14k \\
(2 replicas) & 4   &  33.7 (2.5x) & 10 & 32k\\
(4 replicas) & 8   &{\bf 23.9 (3.5x)} & 14 & 66k 
    & 46 & 7 & 18k \\
 \hline
    Sentiment (82\%) & 1 & 305 & 3 & 88 \\
    & 4 & 230 (1.3x) & 3 & 117 & \\
    & 16 &  {\bf 201 (1.5x)} & 3 & 133
    &  208 & 5 & 265 \\
 \hline
    bAbI 15 (100\%) & 1 & 12.2 & 7 & 319 \\
    & 16 & {\bf 5.8 (2.1x)} &  6 & 662
    & 6.3 & 5 & 350 \\
 \hline
    QM9 (4.6) & 4 & 29k & 93 & 400 \\
    & 16 &  {\bf 14k (2.1x)}  & 69 & 640
    & 129k & 59 &  58\\
   \hline
   \end{tabular}
}  \end{center}\vspace{-2em}
\end{table}
On MNIST, Table \ref{tab:results} shows 3x speedup from synchrony
($\mak=1$) to asynchrony
($\mak=4$). This is almost ideal as the first
three linear layers are the heaviest operations.
As we can see in the fourth column of the table, mild
asynchrony has negligible effect on the convergence while greatly
improving throughput and time to convergence.

The list reduction dataset demonstrates the power of {\em
  replicas}. As there is only one heavy operation (Linear-1, Figure~\ref{fig:rnn-ctrl}),
the speedup from asynchrony is mild (1.3x). However we get 2.5x and 3.5x speedup for 2 and 4
replicas, respectively, which is nearly ideal. Again,
the \# of epochs to convergence is not affected by increasing \mak.
The slowdown in convergence for 4 replicas is due to the increased
{\em effective} minibatch size -- also commonly observed in data parallel
training.

Next the sentiment tree-RNN dataset shows that our runtime is
competitive without batching to TensorFlow Fold \cite{looks2017deep}
using dynamic batching of batch size 100. It is worth mentioning that
our runtime allows us to specify different \texttt{min\_update\_frequency}
parameter for each parameterized operation. We set this parameter to
1000 for the embedding layer, which is initialized by Glove vectors, and
50 for all other layers. This greatly reduced gradient
staleness in the embedding layer.

Finally bAbI 15 (54 nodes) and QM9 datasets demonstrates the importance
of sparsity. Note that the TensorFlow implementation of
GGSNN~\cite{li2015gated} implements the message propagation and
aggregation over the input graph as a dense $NH\times
NH$ matrix multiplication where $N$ is the number of nodes and $H$ is
the hidden state dimension. Since each input graph has a unique
connectivity, this matrix needs to be constructed for each instance. By
contrast, we handle this by message passing and branching as we
described in Section~\ref{sec:ir}. As a result we get roughly 
9x speedup on QM9 against TensorFlow
implementation on CPUs with the same number of threads. Our runtime was
also faster than a GPU TensorFlow implementation by 2.1x. AMPNet and
TensorFlow implementation were comparable on the small bAbI 15 (54
nodes) dataset.

\begin{wrapfigure}{r}{0.48\textwidth}
\vspace*{-15mm}      \includegraphics[width=0.48\textwidth]{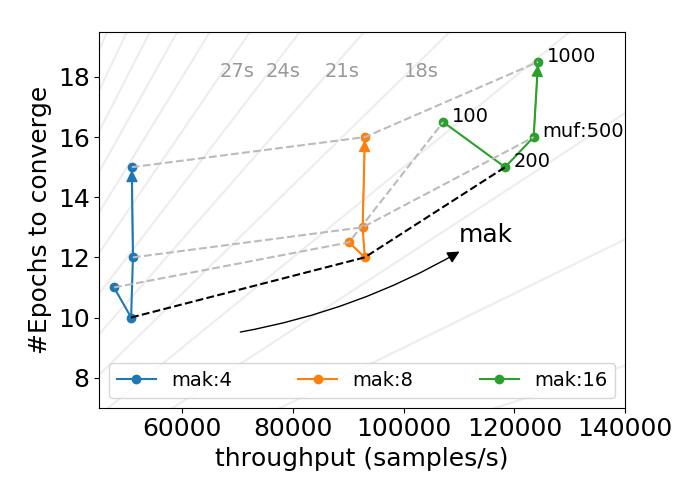}
   \caption{Performance of an 8-replica RNN model on the as a function of asynchrony hyperparameters. Solid gray lines show constant convergence time trajectories. muf stands for {\tt min\_update\_frequency}.\vspace{-1cm}}\label{fig:asynchrony_tuning}
\end{wrapfigure}%

\paragraph{Asynchrony}
The degree of asynchrony is controlled by hyperparameters {\tt
min\_update\_frequency} and \\{\tt max\_active\_keys}.
In Fig. \ref{fig:asynchrony_tuning} we use an 8-replica RNN model on the
list reduction dataset to investigate how these parameters affect the
data and time required to converge to 96\% validation accuracy. We
find, in analogy with minibatch size in traditional systems, that {\tt
min\_update\_frequency} must neither be too large nor too
small. Increasing {\tt max\_active\_keys} (increasing asynchrony)
monotonically increases performance when the number of keys is similar
to the number of individually affinitized heavy operations in the model
8 in this case).
Increasing {\tt max\_active\_keys} significantly
beyond this point produces diminishing returns.

\section{Related Work}
\label{sec:relatedwork}
Chainer \cite{tokui2015chainer},
DyNet \cite{neubig2017dynet}, and PyTorch \cite{pytorch} belong to a new
class of deep learning frameworks that define the computation graph
dynamically per-instance by executing the control flow of the host language (e.g. Python) that can
limit cross-instance parallelism and has a cost that is difficult to hide when the minibatch size is
small\citep[see][]{neubig2017dynet}. By contrast our IR graph is static so it is easier to distribute,
optimize, and pipeline-parallelize across instances.

Theano \cite{team2016theano} and TensorFlow (TF)\cite{abadi2016tensorflow} provide powerful
abstractions for conditional execution (\texttt{ifelse} in Theano and
\texttt{cond} in TF) and loops
(\texttt{scan} and \texttt{while\_loop}, respectively); TF also provides
higher-order functions, such as \texttt{map}, \texttt{foldl}, \texttt{foldr}, and \texttt{scan}.
The main difference between AMPNet and the above frameworks is that
AMPNet is streaming and asynchronous whereas Theano is non-streaming and
synchronous. Although not designed for streaming, TF can support
streaming programmatically as it exposes
first-class queues, as well as data prefetching with so called {\em input pipelines}.
In our IR, all the queuing is implicit and stream-based execution is the default. TF additionally
does support static description of dynamic control flow and state
update, but we depart from the classic dataflow architecture that TF
follows~\cite{Arvind:1986:DA:17814.17824}: First, instead of having
nodes that represent mutable reference cells, we encapsulate the state
with which a message should be processed through the graph in the
message itself. Second, because we encapsulate algorithmic state in the messages, we
do not introduce the notion of control dependencies (which can be used to impose a specific
execution order on TF operations). Our choices complicate algorithmic state management from
a programming point of view and make the task of designing a high-level compiler non-trivial,
but allow every node to run asynchronously and independently without a scheduler and without
the need for control messages: For example, nodes that dynamically take a
control flow path or split the data simply consult the state of the
incoming message, instead of having to accept additional control
inputs. For ``small'' states (e.g. nested loop counters or edge and node
ids) this might be preferable than out-of-band signaling. Our IR can
implement loops by simply using state-update, conditional, and phi
nodes, because the state accompanies the payload throughout its
lifetime, whereas TF introduces specialized operators from timely
dataflow~\cite{Murray:2016:IID:3001840.2983551} to achieve the same effect.

TensorFlow Fold (TFF) \cite{looks2017deep} is a
recent extension of TensorFlow that attempts to increase batching for TF dynamic networks and is
an interesting alternative to our asynchronous execution. TFF unrolls and merges
together (by depth) the computation graphs of several instances, resulting in a batch-like
execution. TFF effectiveness greatly depends on the model -- for example,
it would not batch well for random permutations of a sequence of operations, whereas our IR
would very succinctly express and achieve pipeline parallelism through our control-flow IR nodes.

Asynchronous data parallel training
\cite{recht2011hogwild,dean2012large,chilimbi2014project} is another
popular approach to scale out optimization by removing
synchronization, orthogonal to and combinable with model-parallel training.
For example, convolutional layers are more amenable to data-parallel training than
fully connected layers, because the weights are smaller than the activations.
Moreover, when control flow differs per data instance, data parallelism
is one way to get an effective minibatch size $ > 1$, which may improve convergence
by reducing variance. The impact of staleness on convergence
\cite{recht2011hogwild} and optimization dynamics \cite{mitliagkas2016asynchrony}
have been studied for data parallelism. It would be interesting to extend
those results to our setting.

\citet{jaderberg2016decoupled}, like us, aim to to train different parts of a model in a decoupled or asynchronous
manner. More precisely, their goal is to approximate a gradient
with a {\em synthetic gradient} computed by a small neural network that
is locally attached to each layer. Hence, the local
gradient calculation becomes independent of other layers (except for the
training of the gradient predictor network) and allows asynchronous parameter updates.
This would be especially useful if the evaluation of the local network is cheaper than
the computation of the real gradient; for example, if the computation of the real gradient
required communication of forward/backward messages between devices.

\section{Conclusion and Outlook}
\label{sec:conclusion}
We have presented an asynchronous model-parallel SGD algorithm for
distributed neural network training.
We have described an IR and multi-core CPU runtime
for models with irregular and/or instance-dependent control flow.
Looking forward, we aim to deploy our system on specialized hardware.
To give an idea of performant FPGA implementations of AMPNet,
we perform a simple estimate of the peak throughput on the QM9 dataset
running on a network of 1 TFLOPS FPGAs (see Appendix C for details). Our
calculation shows that we achieve 6k graphs/s (10x compared to our CPU
runtime) on the QM9 dataset with 200 hidden dimensions and 30 nodes per
graph on average. This only requires a very reasonable 1.2 Gb/s network bandwidth.
Equally importantly, we plan to build a compiler that automatically
deduces the information to be placed in the states and generates
state keying functions from a higher-level description of the models.
By unlocking scalable distributed training of dynamic models, we hope to
enable exploration of this class of models that are
currently only on the horizon but may become more mainstream in the future.

\section*{Acknowledgements}
We would like to thank Eric Chung, Doug Burger, and the Catapult team
for continued discussions and feedback from the early stage of our
work. We would also like to thank Krzysztof Jozwik for discussions on
FPGAs, Stavros Volos for discussions on various memory architectures,
Miguel Castro for discussions on data parallelism vs. model parallelism,
John Langford for a discussion on asynchrony and reproducibility, and
Frank Seide for discussions on dynamic networks.

{\small
\bibliography{ampnet}
\bibliographystyle{abbrvnat}
}

\newpage
\appendix

\section{AMPNet runtime implementation}
\label{sec:runtime}


We have implemented an AMPNet runtime for multi-core CPUs.
Our runtime spawns multiple {\em workers} each associated with a hardware thread and hosting one or more IR nodes -- in
  a more general setting each worker corresponds to a compute device. To remain faithful to a distributed environment
communication is only through message passing. Each worker is equipped with a multiple-producer single-consumer queue that can accept messages for any IR node hosted on that worker.

The main worker loop periodically offloads messages from the concurrent queue to a worker-local priority queue that assigns higher priority
to backward messages. Backward prioritization is designed for situations when multiple IR nodes with a dependency on the IR graph end up
hosted on the same worker. As a consequence, backpropagation can complete faster and new instances can be pumped in by the controller.
We dequeue the top message and invoke the forward or backward method of the target IR node. These methods may update internal IR node
state (such as cache the state of the incoming message and wait for more messages) or post new forward or backward messages.


How to update the parameters using the gradients is a configuration option that selects amongst a range of
optimization algorithms. We have implemented runtime configuration options for selecting several well-known schemes
such as (momentum-)SGD and Adam~\cite{kingma2014adam}, and for controlling the training hyper-parameters.

\section{Details of the experimental results}
We provide  more details of the experiment and analysis in this
section. All experiments were carried out on machines with 16 cores and
112 GB of RAM. The validation curves were averaged over at least 20 independent
runs. The time/epoch to reach a target accuracy was calculated as {\em
median} of the time an algorithm takes to reach the target accuracy over
the repetitions. We found this approach to be more reliable than
reporting the time/epoch when the averaged accuracy reaches the
target. Table \ref{tab:throughputs} show both the training and
validation throughputs we obtained with AMPNet and our TensorFlow baselines.

\subsection{MNIST}
\begin{figure}[tb]
  \begin{center}
   \subfigure[MNIST dataset]{
  \includegraphics[width=\textwidth]{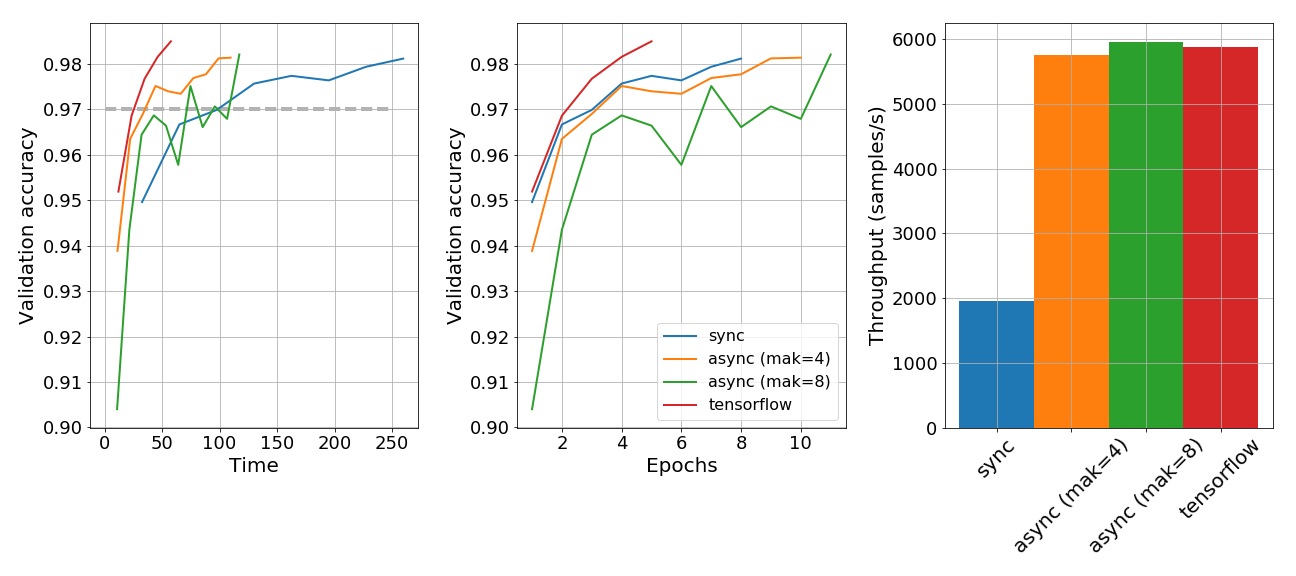}\label{fig:mnist}}
   \subfigure[List reduction dataset]{
  \includegraphics[width=\textwidth]{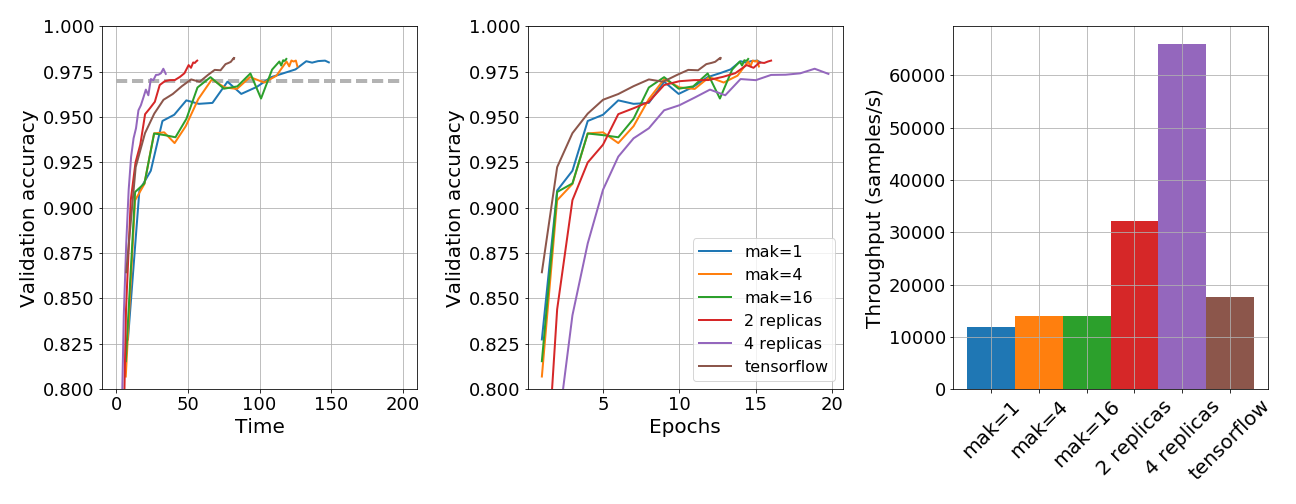}
   \label{fig:calc}}
   \subfigure[Sentiment Tree Bank ($\texttt{min\_update\_frequency}=50$)]{
   \includegraphics[width=\textwidth]{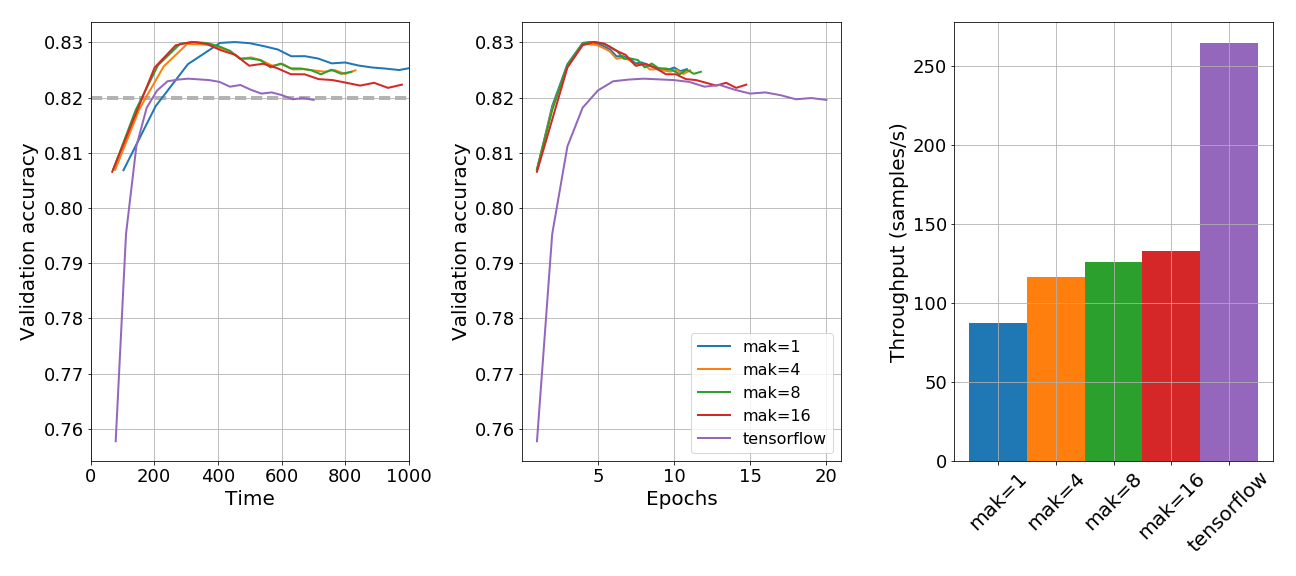}
   \label{fig:sentiment}
   }
  \end{center}
\end{figure}

Figure \ref{fig:mnist} shows the validation accuracy vs. time,
validation accuracy vs. epochs, and throughputs of synchronous and
asynchronous versions of AMPNet as well as TensorFlow. The throughput
greatly increases from synchronous ($\mak=1$) to asynchronous
($\mak=4$) while the speed of convergence (middle
panel) is hardly affected for mild amount of asynchrony. Taking
higher $\mak=8$ increase throughput only very
little (because there is no more work) and seems to rather make the
convergence more unstable. This is due to the fact that our current
scheduler is greedy and pumps in a forward message whenever the first
layer is unoccupied, which leads to large gradient staleness. Clearly a
better scheduling will remove this sensitivity.

\begin{figure}[tb]
  \begin{center}
   \subfigure[Sentiment Tree Bank ($\mak=16$)]{
   \includegraphics[width=\textwidth]{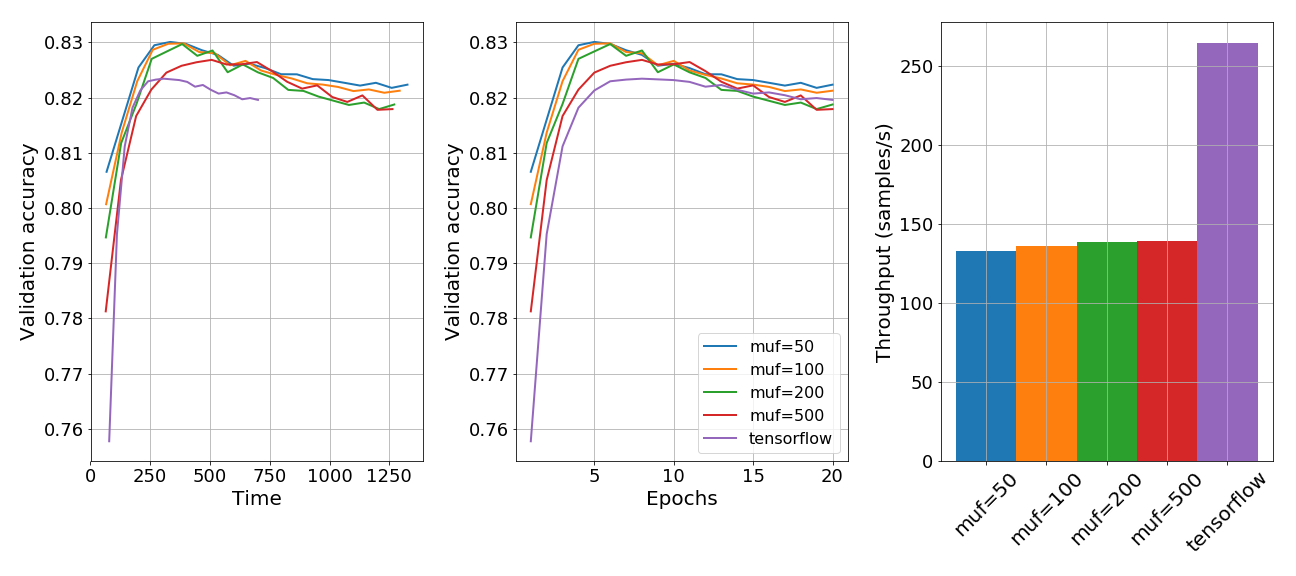}
   \label{fig:sentiment-muf}
   }
  \end{center}
\end{figure}

\begin{figure}[tb]
  \begin{center}
   \subfigure[bAbI 15 (large) dataset]{
  \includegraphics[width=\textwidth]{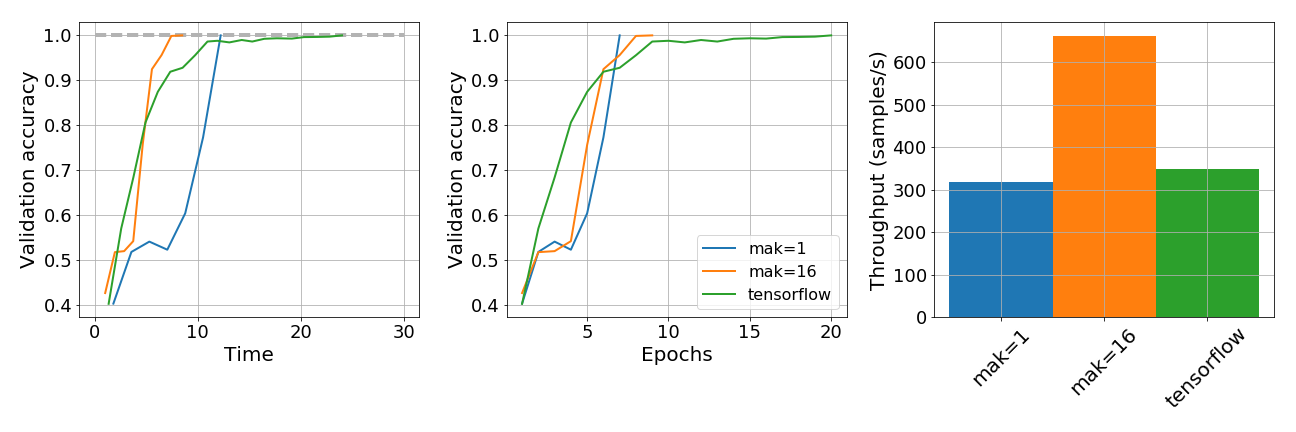}\label{fig:babi}
}
   \subfigure[QM9]{\includegraphics[width=\textwidth]{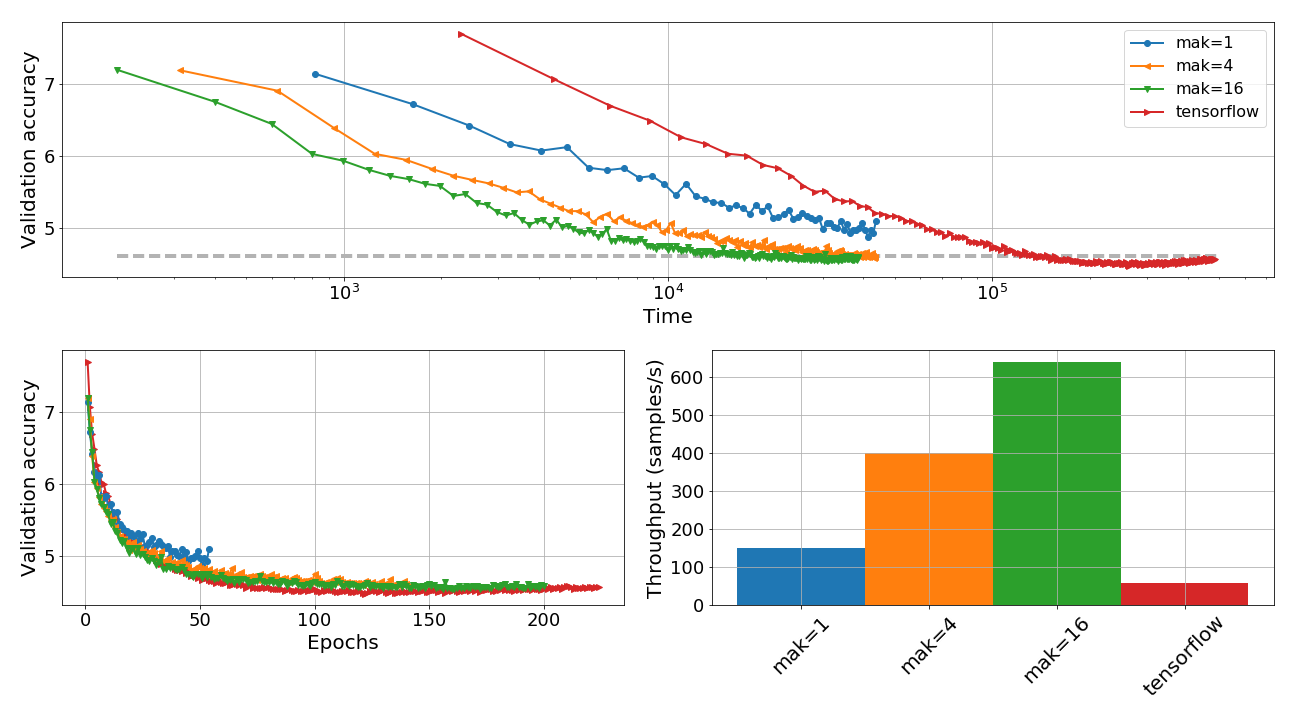}\label{fig:qm9}   }
  \end{center}
 \caption{Convergence plots.}
\end{figure}

\begin{table}[tb]
  \begin{center}
   \caption{Training and validation throughputs. }
   \label{tab:throughputs}
{\small
   \begin{tabular}{@{\extracolsep{4pt}}llllllll@{}}
   \hline
   \hline
    & \multicolumn{2}{l}{number of instances} & \multicolumn{3}{l}{AMP} & \multicolumn{2}{l}{TensorFlow} \\
    \cline{2-3} \cline{4-6} \cline{7-8}
    & train & valid & \texttt{mak} &   train inst/s & valid inst/s &  train inst/s & valid inst/s\\
    \hline
MNIST (97\%) &     &        & 1 &  1948 & 6106 &\\
             & 60k & 10k    & 4 &  5750 & 19113 &  5880 & 8037\\
 \hline
List reduction (97\%) & & & 1  &  12k & 41k \\
                      & & &4  &  14k & 53k\\
                      & & & 16 &  14k & 53k \\
(2 replicas) & & & 4   & 32k & 99k\\
(4 replicas) & 100k& 10k& 8   & 66k &181k & 18k & 43k\\
 \hline
    Sentiment (82\%) & & & 1  & 88 & 326  \\
    & & & 4  & 117 & 568 \\
    & 8511 & 1101 & 16 & 133 & 589 & 265 & 1583\\
 \hline
    bAbI 15 (100\%) & & & 1 &  319 & 733\\
    & 100\footnotemark & 1000 & 16 &  662 & 1428 & 350 & 1093\\
 \hline
    QM9 (4.6) & & & 4  & 400 & 970\\
    & 117k & 13k & 16 &   640 & 1406 &   57.5 & 104\\
   \hline
   \end{tabular}

}  \end{center}
\end{table}
\subsection{List reduction dataset}
Similarly Figure \ref{fig:calc} shows the validation accuracy vs. time
and the number of epochs, and throughputs of the methods we discussed in
the main text on the list reduction dataset. We first notice that
increasing the asynchrony from synchronous (\mak=1) to $\mak=4$ and
$\mak=16$ affects the convergence very little at least in
average. However, there is also very little speedup without introducing
replicas as we discussed in the main text. Increasing the number of
replicas increases the throughput almost linearly from 15k sequences/s
(synchronous) to 30k sequences/s (2 replicas) and over 60k
sequences/s (4 replicas). Convergence is almost unaffected for 2
replicas. This was
rather surprising because the parameters of the replicas are only
synchronized after each epoch as we described in Sec. \ref{sec:replicas}.
A slight slow-down in convergence can be noticed for 4 replicas. Since
even $\mak=16$ has almost no effect on the convergence without replicas,
this is not due to asynchrony. We also tried to synchronize more
frequently but this did not help. Thus we believe that the slow-down is
due to the increase in the effective minibatch size resulting in reduced
number of updates per epoch, which is commonly 
observed in data parallel training.

\subsection{Sentiment Tree Bank dataset}
Figure~\ref{fig:sentiment} shows the averaged fine grained validation accuracy for the tree RNN
model with different
\mak on the Stanford Sentiment Tree Bank dataset. Interestingly
although TensorFlow Fold achieves higher throughput, AMPNet
converges faster (in terms of the number of epochs). This speedup is
mainly due to the fact that we are not batching and updating whenever we
have accumulated 50 gradients (except for the lookup table node that
updates every 1000 gradients); 50 gradients correspond to roughly 2
trees. The reason for the lower throughput compared to TensorFlow Fold
is that we are only grouping the leaf operations and not the branch
operations. Grouping the branch operations is possible by
extending our IR nodes and we are actively working on it.

Figure~\ref{fig:sentiment-muf} shows the same information for
fixed $\mak=16$ and different \texttt{min\_update\_frequency}. We can
see that as we increase \texttt{min\_update\_frequency} from the
originally used 50 to larger values, the peak of the validation accuracy
shifts later and lower becoming closer to the curve obtained by
TensorFlow Fold. This is consistent with the parallels between
\texttt{min\_update\_frequency} and minibatch size we drew in
Section \ref{sec:experiments}.
The \texttt{min\_update\_frequency} parameter has
marginal influence on the training throughput.

\subsection{bAbI 15 (54 nodes) and QM9 dataset}
Figures~\ref{fig:babi} and \ref{fig:qm9} show that GGSNN can tolerate
relatively large $\mak=16$. In particular, on the more challenging QM9
dataset taking $\mak=16$ increased the throughput significantly from 152
graphs/s (synchronous) to 640 graphs/s.

\footnotetext{We sample 100 fresh samples for every epoch.}
\section{Throughput calculation for the GGSNN model for QM9}
Suppose that the hidden dimension $H$ is sufficiently wide so that the
speed of matrix-vector product dominates the throughput of the system
compared to element-wise operations, such as sigmoid and tanh; we take
 $H=200$ in the calculation below.

\begin{figure}[tb]
 \begin{center}
  \includegraphics[width=.7\textwidth]{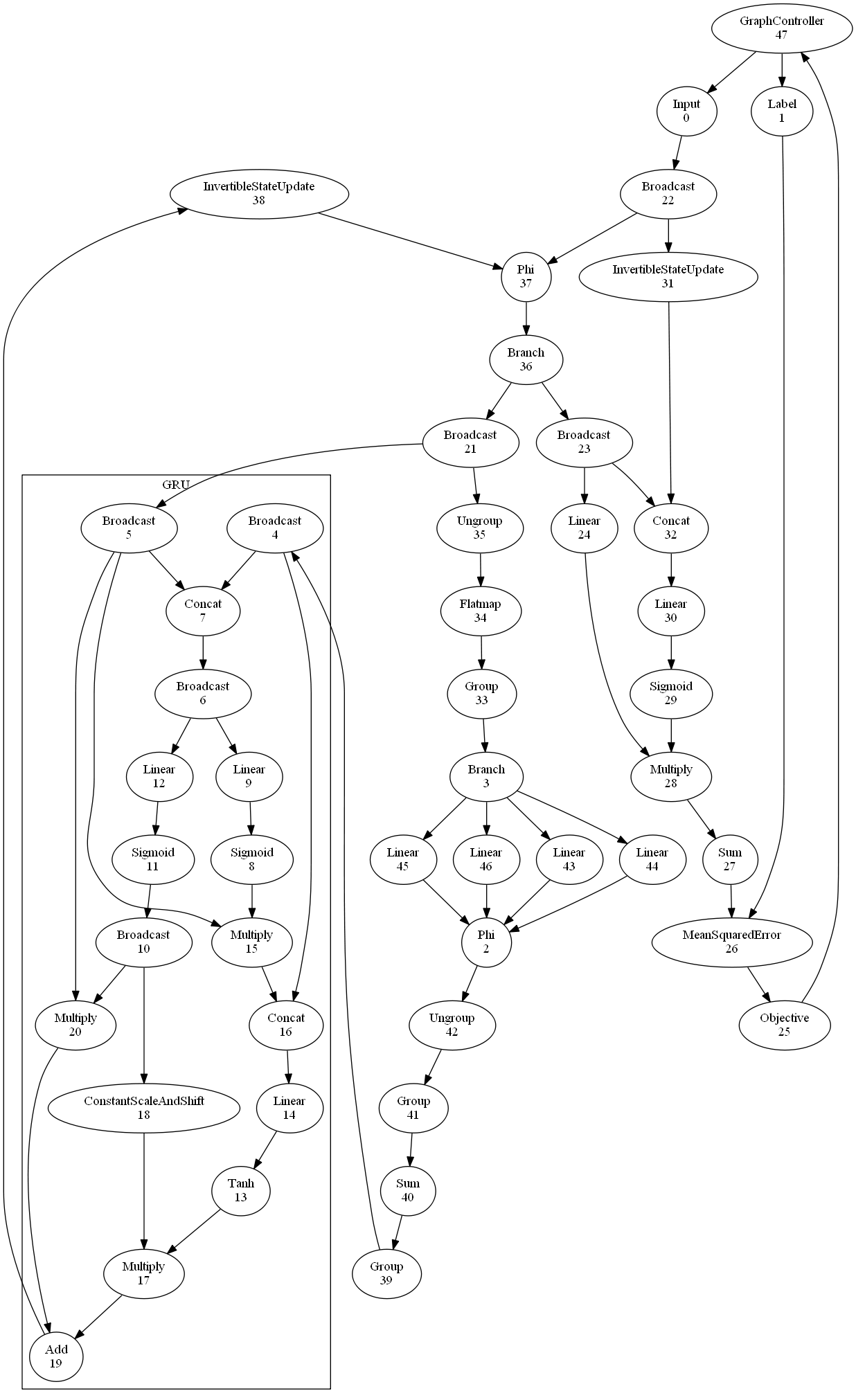}
  \caption{GGSNN IR graph for QM9.}
  \label{fig:molecule-gnn}
 \end{center}
\end{figure}

In an idealized scenario, pipeline parallel execution of the network
consists of roughly 3 stages per time step. In the first
stage, all four $H\times H$ linear nodes corresponding to
different edge types execute in parallel. In the second stage,
the two $2H\times H$ linear nodes (\#9 and \#12) inside
the GRU cell corresponding to update and reset gates execute in
parallel (see Fig.~\ref{fig:molecule-gnn},).
Finally, the last $2H\times H$ linear node in the
GRU cell immediately before the Tanh node executes.
We would need at least 7 devices that executes these linear nodes in a pipelined parallel
fashion. The memory requirement for each device is 4 times the size of
the $H\times H$ or $2H\times H$ weight matrix, which consists of the
parameter, gradient buffer, and two slots for the statistics that need
to be accumulated in the Adam optimizer. This would be 1.2MB for $H=200$
and float32.

The throughput of training this model is either limited by the speed
of the GRU block or that of the linear nodes corresponding to edges.
The number of operations in the forward and backward passes per time
step can thus be estimated as
\begin{align*}
 \textrm{fwdop} &= 2\cdot \max(2NH^2, EH^2/C),\\
 \textrm{bwdop} &= 6\cdot \max(2NH^2, EH^2/C),
\end{align*}
where $N$ and $E$ are the average number of nodes and edges per
instance, respectively and $C$ is the number of edge types, which is $4$
in this task. We assume that the backward operation is 3 times more
expensive than the forward operation because it requires matrix transpose,
matrix multiplication, and gradient accumulation.

Moreover, in an idealized scenario, we can expect that each neural network node
alternates between forward and backward (we thank Vivek Seshadri for
pointing this out).

Thus we can estimate the throughput of training this model on a network
of 1 TFLOPS devices (e.g., Arria 10) as
\begin{align*}
 \textrm{throughput (samples/s)} &= 0.5\cdot
 \frac{10^{12}}{(\textrm{fwdop}+\textrm{bwdop})\cdot 4},
\end{align*}
where the last 4 is the number of propagation time steps and 0.5
accounts for all the other operations and communication overhead  we
ignored in the calculation. 

For $H=200$, $N=E=30$ and $C=4$ we obtain
\begin{align*}
 \textrm{throughput (samples/s)} &= 0.5\cdot \frac{10^{12}}{64\cdot
 NH^2} \simeq 6.5\cdot 10^{3}~\textrm{(samples/s)}.
\end{align*}

 The network bandwidth required in this scenario is
 \begin{align*}
  \textrm{network bandwidth (bits/s)} &= 32\cdot \textrm{throughput}
  \cdot\max(N,E) \cdot H = 1.2\cdot10^{9}~\textrm{(bits/s)}.
 \end{align*}

\end{document}